\documentclass[10pt,twocolumn,letterpaper]{article}

\usepackage{cvpr}              

\usepackage[accsupp]{axessibility}  

\usepackage{graphicx}
\usepackage{amsmath}
\usepackage{amssymb}
\usepackage{booktabs}

\newcommand{\argmin}{\mathop{\rm arg~min}\limits}
\usepackage{mathtools}
\usepackage[toc,page]{appendix}

\usepackage[pagebackref,breaklinks,colorlinks]{hyperref}

\usepackage[capitalize]{cleveref}
\crefname{section}{Sec.}{Secs.}
\Crefname{section}{Section}{Sections}
\Crefname{table}{Table}{Tables}
\crefname{table}{Tab.}{Tabs.}

\def\cvprPaperID{10104} 
\def\confName{CVPR}
\def\confYear{2023}

\begin{document}

\title{Unsupervised Intrinsic Image Decomposition with LiDAR Intensity}

\author{Shogo Sato$^1$, Yasuhiro Yao$^1$, Taiga Yoshida$^1$, \\
Takuhiro Kaneko$^2$, Shingo Ando$^1$, Jun Shimamura$^1$ \\
$^1$NTT Human Informatics Laboratories,
$^2$NTT Communication Science Laboratories\\
{\tt\small \{shogo.sato.wv, taiga.yoshida.ry, jun.shimamura.ec\}@hco.ntt.co.jp}, \\
\tt\small yao-yasuhiro@g.ecc.u-tokyo.ac.jp, \tt\small ando@info.shonan-it.ac.jp
}

\maketitle

\begin{abstract}
  Intrinsic image decomposition (IID) is the task that decomposes a natural image into albedo and shade.
  While IID is typically solved through supervised learning methods, 
  it is not ideal due to the difficulty in observing ground truth albedo and shade in general scenes.
  Conversely, unsupervised learning methods are currently underperforming supervised learning methods
  since there are no criteria for solving the ill-posed problems.
  Recently, light detection and ranging (LiDAR) is widely used 
  due to its ability to make highly precise distance measurements. 
  Thus, we have focused on the utilization of LiDAR, especially LiDAR intensity, to address this issue.
  In this paper, we propose unsupervised intrinsic image decomposition with LiDAR intensity (IID-LI). 
  Since the conventional unsupervised learning methods consist of image-to-image transformations, 
  simply inputting LiDAR intensity is not an effective approach.
  Therefore, we design an intensity consistency loss that computes the error between 
  LiDAR intensity and gray-scaled albedo to provide a criterion for the ill-posed problem.
  In addition, LiDAR intensity is difficult to handle due to its sparsity and occlusion,
  hence, a LiDAR intensity densification module is proposed.
  We verified the estimating quality using our own dataset, which include RGB images, 
  LiDAR intensity and human judged annotations.
  As a result, we achieved an estimation accuracy that outperforms 
  conventional unsupervised learning methods.
\end{abstract}

\section{Introduction}
Intrinsic image decomposition (IID) is the task that aims to decompose a natural image into 
an illumination-invariant component (albedo) and an illumination-variant component (shade), 
and contributes to high level CV tasks such as relighting and scene understanding.
Research on decomposing a natural image has a long history, 
beginning with the proposal of the Retinex theory~\cite{land1971} and IID~\cite{barrow1978}.
Focusing on Lambertian scenes, decomposition of a natural image $I$ is expressed as follows.
\begin{figure}[t]
  \centering
  \includegraphics[width=1.0\linewidth]{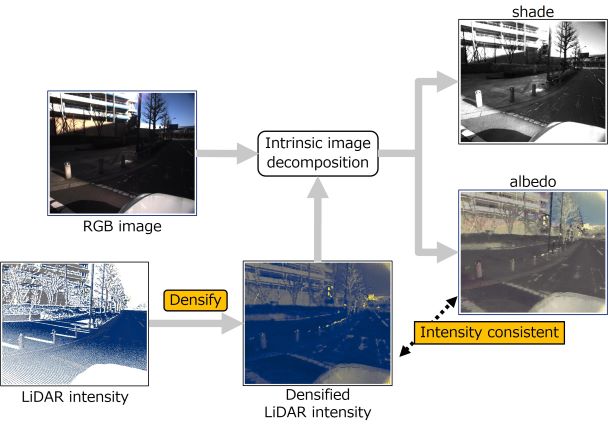}
   \caption{Our proposed approach (IID-LI) is 
   unsupervised intrinsic image decomposition utilizing LiDAR intensity. 
   We densified LiDAR intensity to be robust for LiDAR sparsity or occlusions 
   by a LiDAR intensity densification module.
   In addition, we designed an intensity consistency loss to provide a criterion for the albedo in 
   IID of ill-posed problems.}
  \label{fig1}
\end{figure}
\begin{equation}\label{eq1}
  I=R \cdot S,
\end{equation}
where, $R$ and $S$ denote albedo and shade, respectively.
``$\cdot$'' represents a channel-wise multiplication.
To solve the ill-posed problem,
some researchers assumed that sharp and smooth color variation are 
caused by albedo and shade change, respectively~\cite{land1971,weiss2001,grosse2009,zhao2012}.
As other methods, IID was performed by defining and minimizing energy
based on the assumptions such as albedo flatness~\cite{bell2014,bi2015}.
Moreover, since shades depend on object geometry, 
IID methods with a depth map were also proposed~\cite{chen2013, lee2012, jeon2014}.
With the development of deep learning, supervised learning methods began to be used for 
IID~\cite{narihira2015_1,narihira2015_2,zhou2015,nestmeyer2017,fan2018,luo2020,zhu2021}.
Due to the difficulty of observing ground truth albedo and shade in a practical scenario,
supervised learning methods are typically either small~\cite{grosse2009}, 
synthetic~\cite{butler2012,chang2015,li2018cg} or sparsely annotated~\cite{bell2014}. 
Hence, these supervised learning methods are not ideal for IID in observed data.
To address this issue, a few semi-supervised~\cite{janner2017,yu2019} 
and unsupervised~\cite{ma2018,li2018,liu2020,seo2021} learning methods are proposed.
However, these methods are currently underperforming supervised learning methods 
due to the lack of criteria for solving ill-posed problems by only using image and depth.

In recent years, light detection and ranging (LiDAR), 
which accurately measures the distance to objects, is widely used.
LiDAR usually obtains reflectance intensity (LiDAR intensity) as well as object distance.
Since albedo is the proportion of the incident light and reflected light,
LiDAR intensity utilization as a criterion for albedo helps to solve the ill-posed problem.

In this paper, we propose unsupervised intrinsic image decomposition with LiDAR intensity (IID-LI).
The brief flow of IID-LI is depicted in~\cref{fig1}. 
Since the conventional unsupervised learning methods consist of image-to-image transformations 
based on variational autoencoder (VAE)~\cite{kingma2013} , it is not effective to simply input LiDAR intensity.
Thus, we design an intensity consistency loss that computes the error between 
LiDAR intensity and gray-scaled albedo to provide a criterion 
for the ill-posed problem of decomposing a single image.
In addition, LiDAR intensity is difficult to handle due to its sparsity and occlusion,
hence, LiDAR intensity densification (LID) module is proposed.
The novelty of the LID module lies in the simultaneous convolution of 
sparse data (LiDAR intensity) and dense data of different modality (RGB image). 
Then, we verified the estimating quality with our own dataset 
that combines RGB images and LiDAR intensities in outdoor scenes.
In summary, our contributions are as follows.
\begin{itemize}
  \item We propose LiDAR intensity utilization for intrinsic image decomposition (IID), 
  and an architecture of unsupervised intrinsic image decomposition with LiDAR intensity (IID-LI).
  \item We design an intensity consistency loss to provide 
  a criterion for the ill-posed problem of decomposing a single image.
  \item We propose a LiDAR intensity densification (LID) module based on deep image prior (DIP) 
  to be robust for LiDAR sparsity or occlusions.
  \item We create a publicly available dataset for evaluating IID quality with LiDAR intensity.
  \footnote{Our dataset are publicly available at https://github.com/ntthilab-cv/NTT-intrinsic-dataset}
\end{itemize}
The rest of the paper is organized as follows. 
\cref{sec2} and \cref{sec3} describe related works and baseline methods, respectively.
\cref{sec4} explains our proposed method.
The details of the experiment and experimental results are described in \cref{sec5} and \cref{sec6}, respectively. 
Finally, a summary of the research is given in \cref{sec7}.

\section{Related work}\label{sec2}
In this section, we briefly summarize related works on 
IID and LiDAR intensity utilization.
\subsection{Intrinsic image decomposition (IID)}
\smallskip\textbf{Optimized based methods.}
IID represents an ill-posed problem that decomposes a single image into albedo and shade.
To address this issue, Land et al.~\cite{land1971} proposed a prior that accounts for 
sharp and smooth luminance variations induced by albedo and shade changes, respectively.
Grosse et al.~\cite{grosse2009} improved estimation accuracy 
by considering hue variance as well as luminance.
In addition, several priors have been proposed to enhance the estimation accuracy, 
including the piecewise constant in albedo~\cite{liao2013,barron2014}, 
sparse and discrete values of albedo~\cite{omer2004,rother2011,shen2011}, 
and similar shade among neighboring pixels~\cite{garces2012}.
Bell et al.~\cite{bell2014} formulated and minimized energy function based on these priors.
To achieve edge-preserving smoothing, 
a combination of global and local smoothness based on superpixels was proposed~\cite{bi2015}.
Although these handcrafted priors are reasonable in small images, 
they are insufficient for more complex scenarios, such as outdoor scenes.

\smallskip\textbf{Supervised learning methods.}
With the development of deep learning, supervised learning methods began to be used for IID.
Narihira et al.~\cite{narihira2015_1} first applied supervised learning methods to IID.
In addition, IID was performed by learning 
the relative reflection intensity of each patch extracted from the image~\cite{zhou2015}.
Nestmeyer et al.~\cite{nestmeyer2017} directly estimated per-pixel albedo by trained 
convolutional neural network (CNN).
Fan et al.~\cite{fan2018} designed a loss function for a universal model, 
which works on both fully-labeled and weakly labeled datasets. 
Recently, many researchers have trained supervised learning models based on 
albedo and shade from synthetic data~\cite{luo2020,zhu2021}, 
since observing the ground truth albedo and shade in a practical scenario is difficult. 
However, the estimation accuracy for observed data 
may be limited by the gap between the synthetic and observed data.

\smallskip\textbf{Unsupervised learning methods.}
In these days, semi-supervised and unsupervised learning methods have begun to be used in IID.
Janner et al.~\cite{janner2017} suggested semi-supervised learning methods
that utilize a few labeled data for training and transfer to other unlabeled data.
Most existing unsupervised learning methods such as Li et al.~\cite{li2018} require 
a series of images or multi-view images.
Liu et al.~\cite{liu2020} proposed unsupervised single image intrinsic image decomposition 
(USI$^3$D), which is an unsupervised learning method from a single image.
USI$^3$D outperforms state-of-the-art unsupervised learning methods for IID.
However, these unsupervised learning methods are currently underperforming
supervised learning methods due to the lack of criteria for solving ill-posed problems.
In addition, it is difficult for these methods to discriminate 
between cast shadows and image textures when only using an image and depth.
Therefore, we propose the utilization of LiDAR intensity, 
which is independent of sunlight conditions, cast shadow, and shade.

\smallskip\textbf{Datasets for intrinsic image decomposition.}
MIT Intrinsics, a dataset of image decomposed data on 16 real objects, was published 
by Grosse et al.~\cite{grosse2009} as a dataset for IID.
Since MIT Intrinsics is small for training deep learning models, synthetic data are widely used.
Thus, Butler et al.~\cite{butler2012} collected an MPI Sintel dataset, 
which consisted of synthetic data that included albedo, depth, and optical flow. 
In addition, synthetic datasets 
such as the ShapeNet~\cite{chang2015} and the CGIntrinsics~\cite{li2018cg} were also published.
The free supervision from video games (FSVG) dataset~\cite{krahenbuhl2018}, 
which was extracted from video games and contains a large number of 
images in outdoor settings with albedo. 
On the other hand, Bell et al.~\cite{bell2014} published the IIW dataset in real scenes 
with large number of sparse annotations.
Conventionally, datasets with LiDAR intensity and annotations for IID did not exist. 
To validate the utility of LiDAR intensity for IID, 
we have created a publicly available dataset that includes RGB images, LiDAR intensity, and IID annotations.

\subsection{LiDAR intensity utilization for computer vision}
A LiDAR calculates the object distance from the time lags
between irradiating the laser and detecting the reflected light.
In this process, LiDAR intensity, 
which is the return strength of a irradiated laser beam, is also obtained.
LiDAR intensity is intrinsic to the object surface 
and is therefore independent of sunlight conditions, cast shadow, and shade.
Thus, Guislain et al.~\cite{guislain} performed a shadow detection based
on the un-correlation between color and LiDAR intensity.
In addition, homogeneous regions are extracted with LiDAR intensity and 
elevation for cast shadow detection~\cite{liu2022}.
LiDAR intensity is also utilized for hyper-spectral data correction~\cite{priem2016,brell2017} and 
object recognition~\cite{lang2009,kashani2015,man2015}.
As mentioned above, LiDAR intensity utilization has the potential to 
separate scene illumination from albedo, even in unsupervised manner.
\begin{figure*}[h!]
  \centering
  \includegraphics[width=0.95\linewidth]{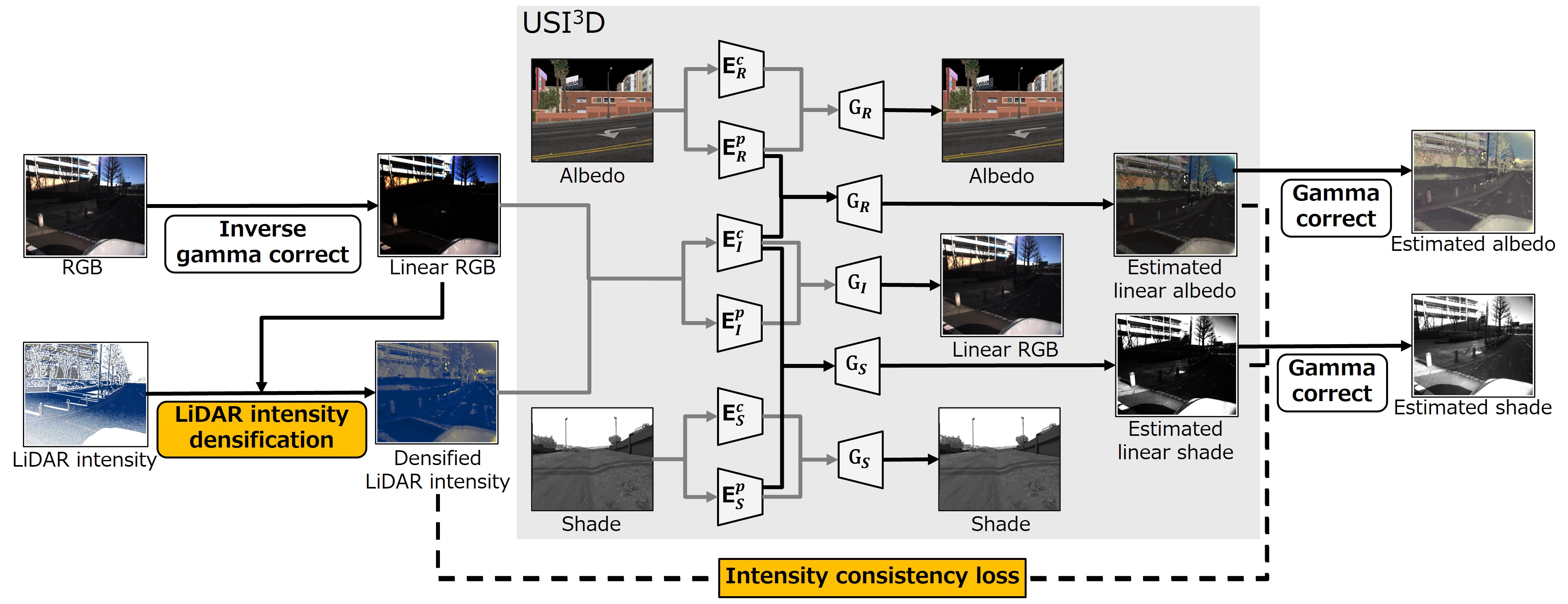}
  \caption{The proposed architecture of IID-LI. 
  Given an RGB image and LiDAR intensity, 
  we convert natural images $I$ into albedo $R$ and shade $S$ domains.
  The input RGB image is transformed by inverse gamma correction 
  to linearize the image values.
  Next, a LiDAR intensity densification module is used for robustness against LiDAR sparsity or occlusions.
  Intensity consistency loss is implemented to provide a criterion for IID.
  The content encoder, prior encoder, and the decoder are denoted by 
  $E_x^c$, $E_x^p$ and $G_x$, respectively.}.
  \label{fig1-}
\end{figure*}

\section{Baseline method}\label{sec3}
In this section, we describe the theory of USI$^3$D~\cite{liu2020}, 
which is the basis of our proposed method.
USI$^3$D was selected as the baseline of IID-LI 
since it is state-of-the-art unsupervised intrinsic image decomposition method.
USI$^3$D~\cite{liu2020} is based on VAE~\cite{kingma2013} 
with generative adversarial net~\cite{goodfellow2020} (VAEGAN)~\cite{larsen2016}.
In USI$^3$D, a sample $X\in$ \{input $I$, albedo $R$, or shade $S$\} 
is decomposed into domain variant ($z_{X}$) and in-variant component ($c_{X}$).
The domain variant component refers to a feature 
that is unique to each of $I$, $R$, and $S$ such as colors, 
while the domain in-variant component is a common feature of them such as edges.
VAE consists of an encoder $E^p_X$ for $z_{X}$ (prior code encoder), 
an encoder $E^c_X$ for $c_{X}$ (content encoder), 
and a generator $G_X$ from $z_{X}$ and $c_{X}$ into a reconstructed image.
The image-to-image transformation from $I$ to estimated albedo $R(I)$ and shade $S(I)$ requires 
the estimation of the domain variant components 
of albedo and shade ($z_R(I), z_S(I)$) corresponding to $I$.
First, the deviation of encoded contents $I$, $R(I)$ and $S(I)$ is calculated.
\begin{equation}\label{eq3}
  \mathcal{L}^{\rm{cnt}}=|c_{R(I)}-c_{I}|+|c_{S(I)}-c_{I}|,
\end{equation}
where $c_{R(I)}$ and $c_{S(I)}$ are the encoded content of $R(I)$ and $S(I)$, respectively.
Second, Kullback-Leibler divergence loss is used to constrain the $z_{R(I)}$ and $z_{S(I)}$ 
in the albedo prior domain $z_{R}$ and shade prior domain $z_{S}$, respectively.
\begin{multline}\label{eq4}
  \mathcal{L}^{\rm{KL}}=\mathbb E[\log p(z_{R(I)})-\log q(z_{R})] \\
  +\mathbb E[\log p(z_{S(I)})-\log q(z_{S})].
\end{multline}
On the other hand, VAE computes $\mathcal{L}^{\rm{img}}$ and $\mathcal{L}^{\rm{pri}}$ 
to reconstruct the input image and prior code, respectively.
\begin{equation}\label{eq5}
  \mathcal{L}^{\rm{img}}=\sum_{x\in I \rm{or} R \rm{or} S}|G_x(E_x^c(x),E_x^p(x))-x|,
\end{equation}
\begin{equation}\label{eq6}
  \mathcal{L}^{\rm{pri}}=\sum_{x\in I \rm{or} R \rm{or} S}|E_x^p(G_x(c_x,z_x))-z_x|.
\end{equation}
In addition, adversarial loss $\mathcal{L}^{\rm{adv}}$ is computed 
so that the generated image fits the target domain.
\begin{multline}\label{eq7}
  \mathcal{L}^{\rm{adv}}_{R}=\log(1-D_{R}(R(I)))+\log(D_R(R))\\
  +\log(1-D_{S}(S(I)))+\log(D_S(S)),
\end{multline}
where $D_R$ and $D_S$ are discriminators for albedo and shade domain. 
The IID is based on physical reconstruction 
and computes the product of albedo and shade to match the input image.
\begin{equation}\label{eq9}
  \mathcal{L}^{\rm{phy}}=|I-R(I)\cdot S(I)|.
\end{equation}
For a piece-wise constant, the smooth loss is calculated as follows,
\begin{equation}\label{eq9+}
  \mathcal{L}^{\rm{smooth}}=
  \sum_{i=1}^{N}\sum_{j\in N(i)}v_{i,j}|\log x_R^i-\log x_R^j|_{1},
\end{equation}
\begin{equation}\label{eq9++}
  v_{i,j}=\exp\left(-\frac{1}{2}(\vec{f}_i-\vec{f}_j)^T\Sigma^{-1}(\vec{f}_i-\vec{f}_j)\right),
\end{equation}
where $N$ and $\vec{f}_i$ are nearest neighbor pixels and $i^{\rm{th}}$ feature vectors, respectively.
The feature vectors are composed of pixel position, image intensity and chromaticity.
USI$^3$D optimizes a weighted sum of the above seven types of losses.
\begin{multline}\label{eq11-}
  \min_{E,G,f}\max_{D}(E,G,f,D)=\mathcal{L}^{\rm{adv}}+\lambda_{1}\mathcal{L}^{\rm{cnt}}
  +\lambda_{2}\mathcal{L}^{\rm{KL}}\\
  +\lambda_{3}\mathcal{L}^{\rm{img}}
  +\lambda_{4}\mathcal{L}^{\rm{pri}}+\lambda_{5}\mathcal{L}^{\rm{phy}}
  +\lambda_{6}\mathcal{L}^{\rm{smooth}}.
\end{multline}

\begin{table*}
  \centering
  \begin{tabular}{@{}ccccc@{}}
    \toprule
    Method & Train Input & Test Input & Learning & Supervised\\
    \midrule
    Baseline R                      & single image & single image & No  & -\\
    Baseline S                      & single image & single image & No  & -\\
    Retinex~\cite{grosse2009}       & single image & single image & No  & -\\
    Color Retinex~\cite{grosse2009} & single image & single image & No  & -\\
    Bell et al.~\cite{bell2014}     & single image & single image & No  & -\\
    Bi et al.~\cite{bi2015}         & single image & single image & No  & -\\
    Revisiting~\cite{fan2018}       & single image & single image & Yes & Yes\\
    IIDWW~\cite{li2018}             & image sequence & single image & Yes & No \\
    UidSequence~\cite{Lettry2018}   & a time-varying image pair & single image & Yes & No \\
    USI$^3$D~\cite{liu2020}         & single image & single image & Yes & No \\
    ours  & single image + LiDAR intensity & single image + LiDAR intensity & Yes & No\\
    \bottomrule
  \end{tabular}
  \caption{\label{table1}A list of comparison methods, their respective data and training categories.}
\end{table*}

\section{Proposed method (IID-LI)}\label{sec4}
\subsection{LiDAR intensity densification module}
When utilizing LiDAR intensity for IID,
the sparsity or occlusion may cause precision degradation.
Thus, we demand to densify the LiDAR intensity before or inside the IID model.
In general, dense LiDAR intensity without occlusion 
corresponding to the supervised data does not exist, 
and the density of LiDAR intensity is sometimes quite low. 
Therefore, it is difficult to letting the network manage it implicitly 
in an end-to-end manner, and we selected a separate model; LID module based on DIP~\cite{ulyanov2018}.
DIP is one of the fully unsupervised learning method used for denoising and inpainting. 
In the original DIP, for an image $x_{0}$ to be cleaned, 
a noise map $z$ is input to a deep neural network $f_{\theta}()$, 
where $\theta$ represents the deep neural network parameters. 
Then, energy optimization is performed to satisfy \cref{eq2}.
\begin{equation}\label{eq2}
  \theta^{*}=\argmin_{\theta}E(f_{\theta}(z);x_{0}).
\end{equation}
Based on the difference in learning speed between the noise component and the image component, 
a clean image is generated by stopping the iteration in the middle of the process.
To apply the original DIP directly to LiDAR intensity densification, 
we substitute LiDAR intensity for $x_{0}$ and optimize \cref{eq2+}.
\begin{equation}\label{eq2+}
  \theta^{*}=\argmin_{\theta}E(f_{\theta}(z);m_{\rm{L}}x_{0}),
\end{equation}
where $m_{\rm{L}}$ is LiDAR mask 
which is 1 for pixels with the observed LiDAR intensity and 0 otherwise.
However, original DIP tend to blur in sparse regions due to its inconsideration of RGB images.
Thus, we input [RGB image, LiDAR intensity]$^T$ and $[m_{\rm{R}},m_{\rm{L}}]^T$ for 
$x_{0}$ and $m_{\rm{L}}$, respectively.
$m_{\rm{R}}$ has the same shape as the RGB image and all pixel values are 1. 
By simultaneous convolution of LiDAR intensity and RGB image,
LiDAR intensity is densified while considering image edges and brightness.
Although DIP is also used in depth completion in multi-view stereo~\cite{zhang2018}, 
the data, objectives, and loss are all different from our model.

\subsection{Intensity consistency loss}
The estimation accuracy of IID is highly dependent on the ability to decompose images 
into domain variant and in-variant components. 
Utilizing LiDAR intensity alone does not adequately enable the network to learn domain dependence. 
Thus, we design an intensity consistency loss that computes the error between LiDAR intensity and 
gray-scaled albedo to provide a criterion for letting the network to learn domain dependence efficiently.
The first term is the loss function when LiDAR intensity $L$ and 
the luminance of the estimated albedo $F(R(I))$ are correlated, 
where $F()$ is the function to convert from RGB image to gray scale. 
In the case where LiDAR intensity and albedo are correlated, 
the image divided by LiDAR intensity is correlated to the shade from~\cref{eq1}.
Thus, the second term represents the loss function when $F(I)/L$ and 
the luminance of the estimated shade $F(S(I))$ are correlated. 
\begin{multline}\label{eq10}
  \mathcal{L}^{\rm{int}}=|F(R(I))-s_{1}L-b_{1}|\cdot m_{\rm{L}} \\
  + |F(S(I))-s_{2}(F(I)/L)-b_{2}|\cdot m_{\rm{L}},
\end{multline}
where $m_{\rm{L}}$ is the mask, which is 1 for pixels with densified LiDAR intensity and 0 otherwise.
In addition, $s_x$ and $b_x$, where $x\in {1,2}$, 
are trainable parameters for adjusting the scale and bias of LiDAR intensity, respectively.
In summary, IID-LI optimizes the loss function in \cref{eq11}.
\begin{multline}\label{eq11}
  \min_{E,G,f}\max_{D}(E,G,f,D)=\mathcal{L}^{\rm{adv}}+\lambda_{1}\mathcal{L}^{\rm{cnt}}
  +\lambda_{2}\mathcal{L}^{\rm{KL}}\\
  +\lambda_{3}\mathcal{L}^{\rm{img}}
  +\lambda_{4}\mathcal{L}^{\rm{pri}}+\lambda_{5}\mathcal{L}^{\rm{phy}}
  +\lambda_{6}\mathcal{L}^{\rm{smooth}}+\lambda_{7}\mathcal{L}^{\rm{int}}
\end{multline}
As with the original paper~\cite{liu2020}, we set 
$\lambda_{1}, \lambda_{2}, \lambda_{3}, \lambda_{4}, \lambda_{5}, \lambda_{6}$ as 
10.0, 0.1, 10.0, 0.1, 5.0, and 1.0, respectively.
In addition, we set $\lambda_{7}$ as 20.0 in this paper.

\subsection{Network architecture}
The overview of our proposed method is depicted in \cref{fig1-}.
First, LiDAR intensity is densified by LID module 
since LiDAR points are usually sparse and have occlusion.
Then, a RGB image and a densified-LiDAR intensity are given to 
USI$^3$D-based model with the intensity consistency loss.
Since images are generally gamma-corrected for human visibility, 
the observed brightness is converted non-linearly.
Thus, to compare the images linearly with LiDAR intensity, 
an inverse gamma correction was performed before input to IID-LI, 
and gamma correction was performed on the output images.

\begin{table*}
  \centering
  \begin{tabular}{@{}cccccc@{}}
    \toprule
    Method & trained dataset & WHDR & precision & recall & F-score \\
    \midrule
    Baseline R                      &-&0.531&0.393&0.445&0.306 \\
    Baseline S                      &-&0.185&0.431&0.340&0.314 \\
    Retinex~\cite{grosse2009}       &-&0.187&0.496&0.455&0.469 \\
    Color Retinex~\cite{grosse2009} &-&0.187&0.496&0.455&0.470 \\
    Bell et al.~\cite{bell2014}     &-&0.213&0.467&0.463&0.457 \\
    Bi et al.~\cite{bi2015}         &-&0.283&0.462&0.522&0.466 \\
    Revisiting~\cite{fan2018} & IIW &\textbf{0.181}&\textbf{0.575}&0.485&0.499 \\
    IIDWW~\cite{li2018}     & BIGTIME &0.375&0.418&0.483&0.397\\
    UidSequence~\cite{Lettry2018} & SUNCG-II &0.372&0.405&0.453&0.395 \\
    USI$^3$D~\cite{liu2020}& IIW + CGIntrinsics &0.347&0.428&0.497&0.418 \\
    USI$^3$D (ours)~\cite{liu2020}& ours + FSVG   &0.287&0.444&0.504&0.446 \\
    Ours (without LID) & ours + FSVG   &0.283&0.459&0.530&0.467\\
    Ours (without $\mathcal{L}^{\rm{int}}$) & ours + FSVG   &0.330&0.426&0.483&0.421\\
    Ours               & ours + FSVG   &0.227&0.517&\textbf{0.591}&\textbf{0.521}\\
    \bottomrule
  \end{tabular}
  \caption{\label{table2}Numerical comparison with our dataset for 
  \textbf{all} (E=9411, D=2554, L=661) annotation points.}
  \end{table*}

\section{Experimental setups}\label{sec5}
\subsection{Data preparation}\label{sec5-1}
In this paper, we attempt to utilize LiDAR intensity for IID.
However, no public data with both LiDAR data and annotations for IID exists.
Therefore, we employed our own dataset consisting of images, LiDAR data, and annotation.
The data was collected using a mobile mapping system (MMS) equipped with RGB camera, LiDAR, 
global navigation satellite system (GNSS) and an inertial measurement unit (IMU).
For LiDAR scanning, the ZF profiler which features 0.0009 degree angular resolution, 
1.02 million points per second, 360-degree field of view and a maximum range of 120 meters.
The data was recorded at Shinagawa, Yokohama, and Mitaka in Japan.

\subsection{Annotation}\label{sec5-2}
The annotation in this study follows the conventional method~\cite{bell2014}.
First, we extracted 100 samples from the obtained dataset for sparse annotation.
We then utilized Poisson disk sampling with a minimum radius of 7\% of image size to sample image points, 
and the points with over and under-saturation or those around the edge 
were removed in the same manner as Bell et al~\cite{bell2014}.
Finally, Delaunay triangulation was performed to generate edges, 
resulting in $91 \pm 24$ pairs per image as sparse sampling.
Dense sampling data was generated using Poisson disk sampling 
with a minimum radius 3\% of image size, resulting in $566 \pm 143$ pairs per image.

In this study, 10 annotators, who understand the concept of albedo, performed the annotation. 
For each pair, 5 out of 10 annotators answered the following three questions.
\begin{itemize}
  \item Do the two points have the same albedo intensity?
  \item If not, does the darker point have a darker surface albedo intensity?
  \item How confident (Definitely, Probably, Guessing) are you in your judgment of the above?
\end{itemize}
$a_{i,j}$ is the $i^{\rm{th}}$ annotator judgement for $j^{\rm{th}}$ question.
The $a_{i,1}$ and $a_{i,2}$ value +1 for "yes" and -1 for "no" 
for the first and second questions, respectively.
The $a_{i,3}$ values 1.0, 0.8 and 0.3 for "Definitely", "Probably" and "Guessing", respectively.
In this study, the judgement $j$ and the weight $w$ for each pair were calculated as following:
\begin{equation}\label{eq12}
  (J, w)= \left\{ 
  \begin{aligned} 
    (E, A_{1}) &\ \ \rm{if}\ \ A_{1}>0\\
    (D, A_{2}) &\ \ \rm{if}\ \ A_{1}\leq0 \ A_{2}>0\\
    (L, -A_{2})&\ \ \rm{else},
  \end{aligned}
  \right.
\end{equation}
where, $A_{1}=\sum\limits_{i=1}^5 a_{i,1}a_{i,3}$ and 
$A_{2}=\sum\limits_{i=1}^5 a_{i,2}a_{i,3}$.
Note that, annotations are utilized only for quantitative assessment of estimation accuracy.

\subsection{Quantitative evaluation metric}\label{sec5-3}
For evaluating the estimation accuracy, 
we defined the threshold differences between the points used in the human judgements.
\begin{equation}\label{eq13}
  (\hat J)= \left\{ 
  \begin{aligned}
    D &\ \ \rm{if}\ \ R_L/R_D>1+\delta\\
    L &\ \ \rm{if}\ \ R_D/R_L>1+\delta\\
    E &\ \ \rm{else},
  \end{aligned}
  \right.
\end{equation}
where, $R_L$ and $R_D$ are lighter and darker points of a annotated points pair.
Following the conventional studies~\cite{bell2014}, we set the threshold $\delta=0.1$. 
We use four indices: 
weighted human disagreement rate (WHDR), precision, recall, and F-score.
WHDR is calculated as \cref{eq14}.
\begin{equation}\label{eq14}
  \rm{WHDR}_{\delta}(J,R)=
  \frac{\sum_{k} w_k \cdot \mathbf{1}(J_k \neq \hat J_{k,\delta}(R))}{\sum_{k}w_k}.
\end{equation}
The human judgement weights are also taken into account 
when computing precision, recall, and F-score.

\subsection{Training details}\label{sec5-4}
Our prepared dataset consisted of 10000 samples with RGB images and LiDAR intensities for training.
Moreover, we prepared an additional 110 samples for testing, which were combined with a total of 12,626 human judgments.
Since IID-LI is based on VAEGAN, both albedo and shade domain data groups are required. 
Furthermore, to make training fair, the albedo and shade datasets were independent each other.
Thus, the albedo dataset was created by extracting 10000 albedo samples from the FSVG dataset~\cite{krahenbuhl2018}. 
The shade dataset was created with no overlap with the albedo dataset.

\begin{figure*}[h!]
  \centering
  \includegraphics[width=1.0\linewidth]{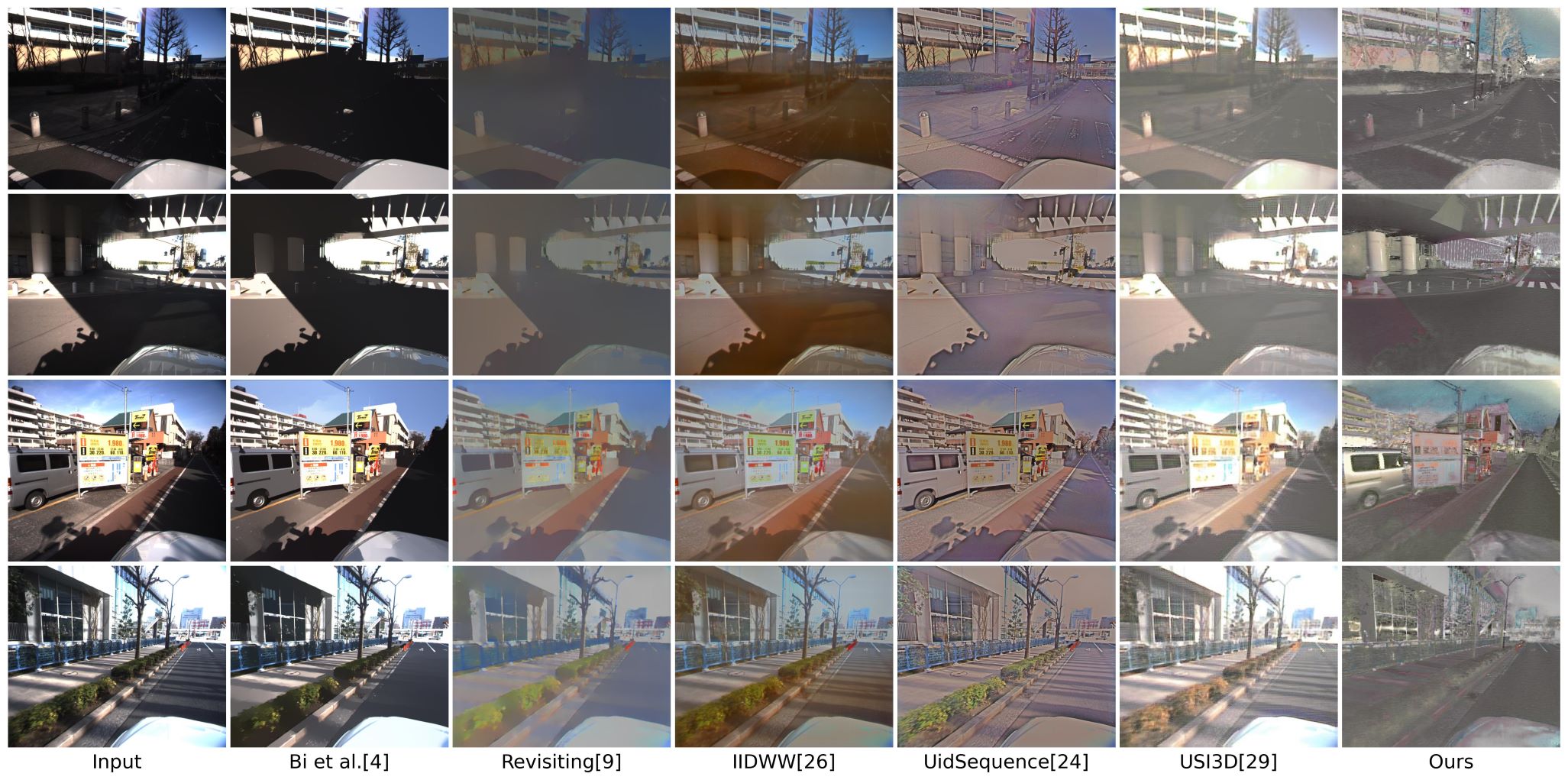}
  \caption{Four examples for IID-LI and compared methods including 
  Bi et al.~\cite{bi2015}, Revisiting~\cite{fan2018}, IIDWW~\cite{li2018}, 
  UidSequence~\cite{Lettry2018} and USI$^3$D~\cite{liu2020}. 
  The conventional methods exhibited noticeable cast shadows, which were reduced with the proposed method.}
  \label{fig2}
\end{figure*}

\section{Evaluation}\label{sec6}
Initially, we present ten conventional methods for comparison. 
Next, we provide a quantitative evaluation of these methods 
based on the metrics of WHDR, precision recall, and F-score.

\subsection{Compared methods}\label{sec6-1}
In this section, we selected ten compared methods including 
both optimized-based and learning-based methods listed in \cref{table1}. 
Firstly, "Baseline R" represents an image where all pixel values are 1.
Conversely, "Baseline S" decomposes the input image with all the shade pixel values as 1.
In addition, Retinex~\cite{grosse2009}, Color Retinex~\cite{grosse2009}, 
Bell et al.~\cite{bell2014}, and Bi et al.~\cite{bi2015} 
were selected for optimized-based methods. 
Finally, supervised learning methods such as Revisiting~\cite{fan2018}, 
and unsupervised learning methods such as IIDWW~\cite{li2018}, UidSequence~\cite{Lettry2018}, 
and USI$^3$D~\cite{liu2020} were employed. 
For all learning-based methods, publicly available parameters 
and pre-trained models were used as defaults.
Since USI$^3$D is the baseline method, we retrained and validated it with our dataset.

\begin{table}
  \centering
  \begin{tabular}{@{}cccccc@{}}
    \toprule
    Method & WHDR & precision & recall & F-score \\
    \midrule
    Baseline R                      &0.527&0.375&0.440&0.35 \\
    Baseline S                      &0.529&0.361&0.340&0.227 \\
    Retinex~\cite{grosse2009}       &0.452&0.523&0.445&0.42 \\
    Color Retinex~\cite{grosse2009} &0.452&0.531&0.445&0.42 \\
    Bell et al.~\cite{bell2014}     &0.446&0.504&0.453&0.414 \\
    Bi et al.~\cite{bi2015}         &0.406&0.561&0.522&0.49 \\
    Revisiting~\cite{fan2018}    &0.428&\textbf{0.635}&0.470&0.442 \\
    IIDWW~\cite{li2018}      &0.464&0.489&0.475&0.417\\
    UidSequence~\cite{Lettry2018}  &0.483&0.453&0.450&0.419 \\
    USI$^3$D~\cite{liu2020} &0.432&0.534&0.500&0.451 \\
    USI$^3$D (ours)~\cite{liu2020}  &0.422&0.539&0.500&0.454 \\
    Ours (without LID)   &0.410&0.547&0.532&0.534\\
    Ours (without $\mathcal{L}^{\rm{int}}$)   &0.455&0.513&0.473&0.430\\
    Ours & \textbf{0.353}&0.625&\textbf{0.596}&\textbf{0.602}\\
    \bottomrule
  \end{tabular}
  \caption{\label{table3}Numerical comparison with our dataset for 
  \textbf{randomly sampled} (E=661, D=661, L=661) annotation points,
  to eliminate bias in the number of annotations.}
\end{table}

\subsection{Qualitative and quantitative evaluation}\label{sec6-2}
First, we quantitatively evaluated our proposed method and the compared methods.
As shown in \cref{table2}, our proposed method achieved 
the state-of-the-art in terms of recall and F-score.
However, Revisiting~\cite{fan2018} was the best for WHDR and precision.
The estimation accuracy of each model may not have been properly evaluated
due to the significant bias in the number of ``E", ``D" and ``L" annotations (E=9411, D=2554, L=661).
Thus, to eliminate bias in the number of annotations, 
we randomly sampled the annotations so that each annotation size is the same (E=661, D=661, L=661).
\cref{table3} shows the evaluating result for randomly sampled annotations.
Our proposed method outperformed other unsupervised learning methods, 
as shown in~\cref{table2} and \cref{table3}. 
Furthermore, it delivered comparable performance to Revisiting~\cite{fan2018}, 
which is a supervised learning methods.
The visual results are shown in~\cref{fig2}.
Cast shadows remained in conventional methods since these methods 
cannot differentiate between cast shadows and textures. 
On the other hand, the cast shadows were less noticeable in our proposed method by using LiDAR intensity.
However, the images generated by our method are slightly blurred.
The image blur is considered to be caused by the calibration errors 
between the image pixel and LiDAR point, 
thus addressing this issue is a future work.

\begin{table}
  \centering
  \begin{tabular}{@{}ccccc@{}}
    \toprule
  Density & WHDR & precision & recall & F-score \\
  \midrule
  all   &0.353&0.625&0.596&0.602 \\
  50\%  &0.397&0.574&0.567&0.569 \\
  10\%  &0.437&0.532&0.521&0.524 \\
  1\%   &0.481&0.485&0.480&0.481 \\
  \bottomrule
  \end{tabular}
\caption{\label{table4}
Ablation study for LiDAR sparsity \textbf{with} LID module,
when the density of LiDAR intensity was reduced to 50\%, 10\% and 1\% from the original LiDAR intensity.}
\end{table}

\begin{table}
  \centering
  \begin{tabular}{@{}ccccc@{}}
    \toprule
    Density & WHDR & precision & recall & F-score \\
    \midrule
    all   &0.410&0.547&0.532&0.534 \\
    50\%  &0.413&0.541&0.524&0.524 \\
    10\%  &0.461&0.493&0.467&0.434 \\
    1\%   &0.480&0.445&0.453&0.418 \\
    \bottomrule
  \end{tabular}
  \caption{\label{table5}
  Ablation study for LiDAR sparsity \textbf{without} LID module,
  when the density of LiDAR intensity was reduced to 50\%, 10\% and 1\% from the original LiDAR intensity.}
  \end{table}

\begin{figure}
    \begin{minipage}[]{.24\linewidth}
      \centering
       \includegraphics[keepaspectratio,width=.99\linewidth]{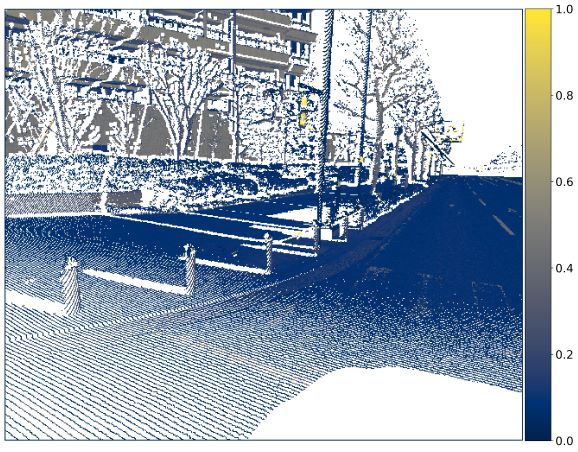}
       \subcaption{all w/o LID}
     \end{minipage}
     \begin{minipage}[]{.24\linewidth}
       \centering
       \includegraphics[keepaspectratio,width=.99\linewidth]{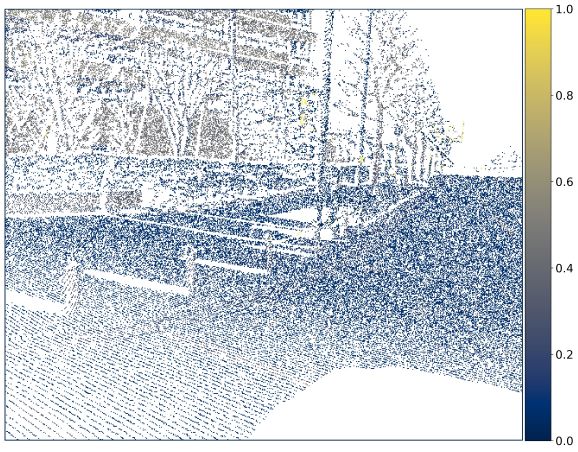}
       \subcaption{50\% w/o LID}
     \end{minipage}
     \begin{minipage}[]{.24\linewidth}
       \centering
       \includegraphics[keepaspectratio,width=.99\linewidth]{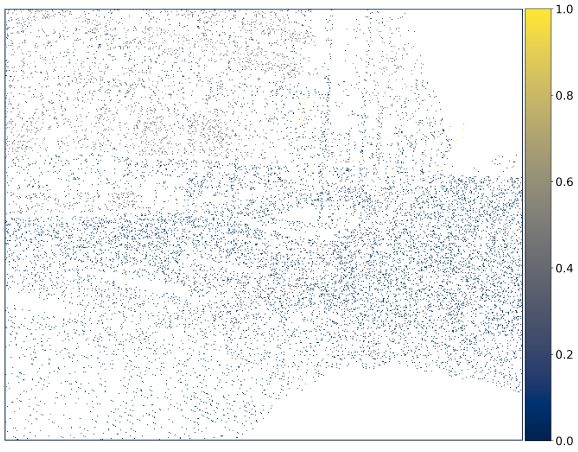}
       \subcaption{10\% w/o LID}
     \end{minipage}
     \begin{minipage}[]{.24\linewidth}
       \centering
       \includegraphics[keepaspectratio,width=.99\linewidth]{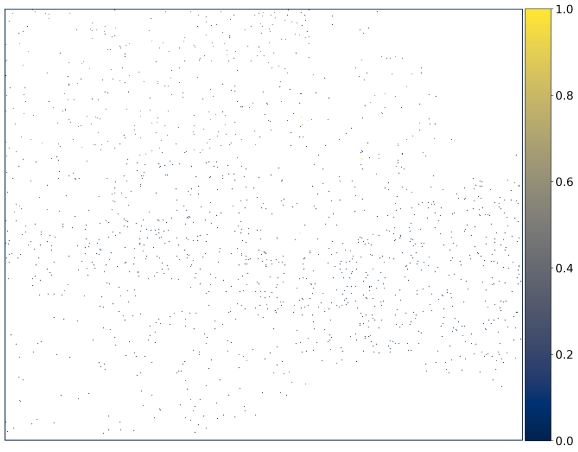}
       \subcaption{1\% w/o LID}
     \end{minipage}\\
     \begin{minipage}[]{.24\linewidth}
       \centering
       \includegraphics[keepaspectratio,width=.99\linewidth]{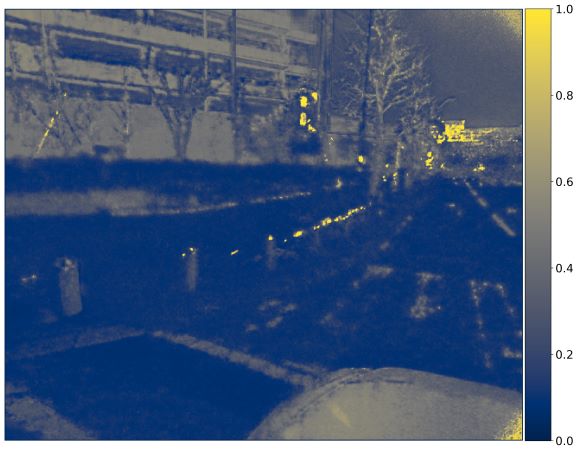}
       \subcaption{all w/ LID}
     \end{minipage}
     \begin{minipage}[]{.24\linewidth}
       \centering
       \includegraphics[keepaspectratio,width=.99\linewidth]{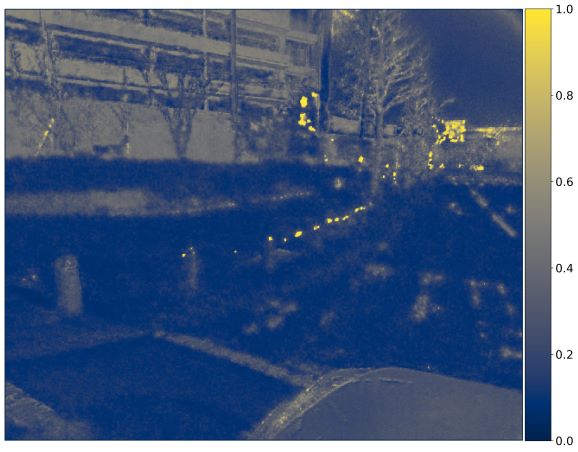}
       \subcaption{50\% w/ LID}
     \end{minipage}
     \begin{minipage}[]{.24\linewidth}
       \centering
       \includegraphics[keepaspectratio,width=.99\linewidth]{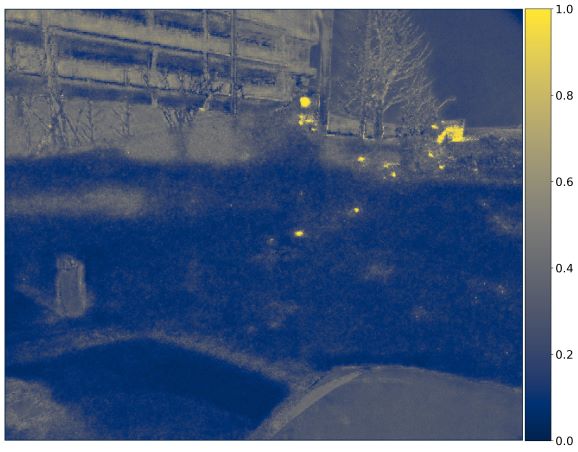}
       \subcaption{10\% w/ LID}
     \end{minipage}
     \begin{minipage}[]{.24\linewidth}
       \centering
       \includegraphics[keepaspectratio,width=.99\linewidth]{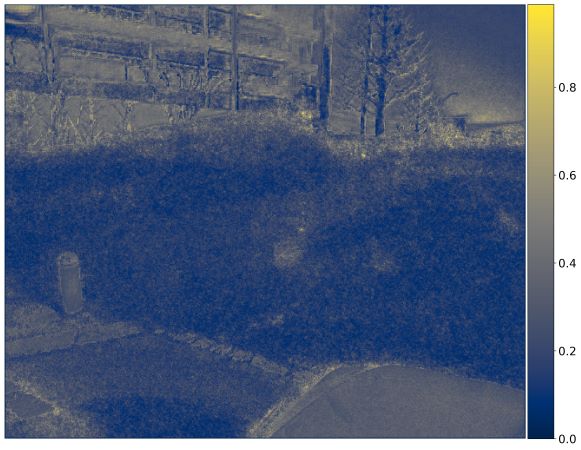}
       \subcaption{1\% w/ LID}
     \end{minipage}
     \caption{An Example of LiDAR intensity with reduced density and the impact of an LID module.
     (a)-(d): Results of sampling LiDAR points to each density. 
     From left to right, LiDAR points were sampled at 100\%, 50\%, 10\%, and 1\%.
     (e)-(h): Result of densifying LiDAR intensity at each density with LID module.}
     \label{fig3}
\end{figure}

\subsection{Ablation study}\label{sec6-3}
In our proposed method, we  have presented LID module and intensity consistency loss. 
Thus, the efficacy of these components has been quantified in \cref{table2} and \cref{table3}.
From the F-score in \cref{table3}, the intensity consistency loss contributed the most to estimation accuracy.

This study used relatively high-density LiDAR data, though it can be sparse in certain scenarios. 
Therefore, we evaluated the estimation accuracy with reduced LiDAR data, 
and the numerical results for random sampled annotations are listed in \cref{table4}.
The density of LiDAR intensity was reduced to 50\%, 10\% and 1\% from the original LiDAR intensity.
As expected, the higher density of LiDAR intensities indicates higher estimation accuracy, as shown in~\cref{table4}.
Specifically, LiDAR intensity reduced to 1\% achieve comparable F-score to the base model, USI$^3$D~\cite{liu2020}.
As a reference, \cref{table5} shows the estimated results without the LID module, 
and indicate that the proposed LID module improves the estimation accuracy.
\cref{fig3} shows an example of LiDAR intensity with reduced LiDAR density, and the effect of LID module.
The LiDAR intensity with 1\% points were densified successfully in visual, 
though the estimation accuracy may be limited due to its blurred detail.
In a future perspective, the development of a more sparsity-robust model will be interesting.

\subsection{Correspondence between albedo and LiDAR intensity}
\begin{figure}[h!]
  \begin{minipage}[]{.32\linewidth}
    \centering
    \includegraphics[keepaspectratio,width=.99\linewidth]{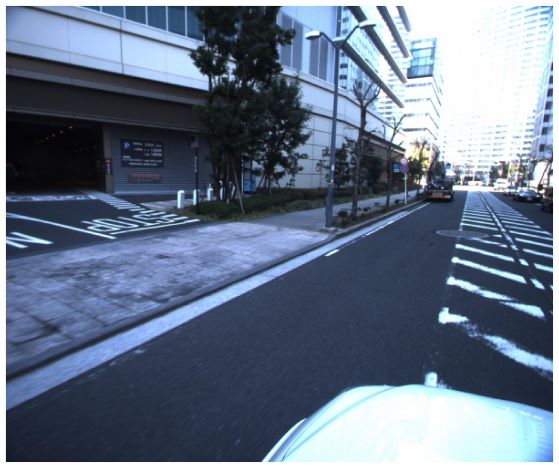}
    \subcaption{RGB image}
  \end{minipage}
  \begin{minipage}[]{.32\linewidth}
    \centering
    \includegraphics[keepaspectratio,width=.99\linewidth]{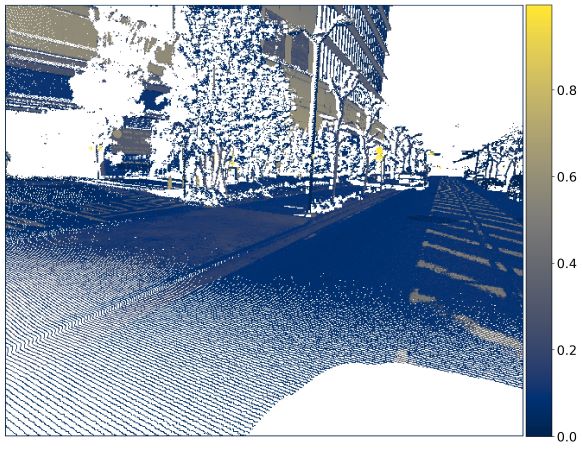}
    \subcaption{LiDAR intensity}
  \end{minipage}
  \begin{minipage}[]{.32\linewidth}
    \centering
    \includegraphics[keepaspectratio,width=.99\linewidth]{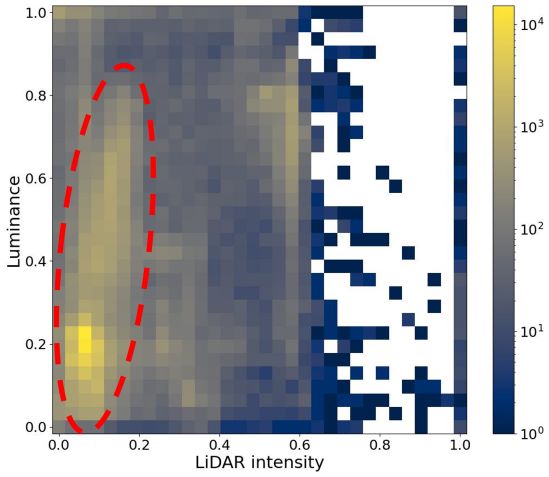}
    \subcaption{Correspondence}
  \end{minipage}
  \caption{Examples of (a) an image in which the field of view is 
  mostly in shadow and (b) the corresponding LiDAR intensities. 
  (c) The correspondence between the luminance and LiDAR intensity are also shown.
  The red dotted lines in (c) indicate areas of high correlation.
  The correlation coefficient values 0.45.}
  \label{fig4}
\end{figure}

The intensity consistency loss computes the error between LiDAR intensity and gray-scaled albedo.
The validity of the intensity consistency loss is discussed in this section. 
First, an image in which the field of view is almost in shadow, 
and the corresponding LiDAR intensity is shown in \cref{fig4}. 
In addition, each corresponding pixel filled into a 2-dimensional histogram is also shown in \cref{fig4} (right).
The correlation coefficient between luminance and LiDAR intensity values for this sample was 0.45.
Since these maps diverge slightly, we expect that the actual correlations coefficient would be a bit higher.
This intensity consistency loss is considered to be effective at least 
in outdoor scenes with mostly achromatic materials such as concrete and wall surfaces.

\section{Conclusion}\label{sec7}
In this paper, we proposed unsupervised intrinsic image decomposition with LiDAR intensity. 
We designed an intensity consistency loss and added an LID module for effective LiDAR intensity utilization. 
As a result, our method outperforms the conventional 
unsupervised learning methods and is comparable to supervised learning methods with our dataset.
In future perspective, we will improve the robustness of LiDAR sparsity, 
and the registration of image and LiDAR intensity.

{\small
\bibliographystyle{ieee_fullname}
\bibliography{main}

\begin{thebibliography}{10}\itemsep=-1pt

\bibitem{barron2014}
Jonathan~T Barron and Jitendra Malik.
\newblock {Shape, illumination, and reflectance from shading}.
\newblock {\em IEEE TPAMI}, 37(8):1670--1687, 2014.

\bibitem{barrow1978}
Harry Barrow, J Tenenbaum, A Hanson, and E Riseman.
\newblock {Recovering intrinsic scene characteristics}.
\newblock {\em Computer Vision Systems}, 2(3-26):2, 1978.

\bibitem{bell2014}
Sean Bell, Kavita Bala, and Noah Snavely.
\newblock {Intrinsic images in the wild}.
\newblock {\em ACM TOG}, 33(4):1--12, 2014.

\bibitem{bi2015}
Sai Bi, Xiaoguang Han, and Yizhou Yu.
\newblock {An l 1 image transform for edge-preserving smoothing and scene-level
  intrinsic decomposition}.
\newblock {\em ACM TOG}, 34(4):1--12, 2015.

\bibitem{brell2017}
Maximilian Brell, Karl Segl, Luis Guanter, and Bodo Bookhagen.
\newblock {Hyperspectral and lidar intensity data fusion: A framework for the
  rigorous correction of illumination, anisotropic effects, and cross
  calibration}.
\newblock {\em IEEE Transactions on Geoscience and Remote Sensing},
  55(5):2799--2810, 2017.

\bibitem{butler2012}
Daniel~J Butler, Jonas Wulff, Garrett~B Stanley, and Michael~J Black.
\newblock {A naturalistic open source movie for optical flow evaluation}.
\newblock In {\em ECCV}, pages 611--625. Springer, 2012.

\bibitem{chang2015}
Angel~X Chang, Thomas Funkhouser, Leonidas Guibas, Pat Hanrahan, Qixing Huang,
  Zimo Li, Silvio Savarese, Manolis Savva, Shuran Song, Hao Su, et~al.
\newblock {Shapenet: An information-rich 3d model repository}.
\newblock {\em arXiv preprint arXiv:1512.03012}, 2015.

\bibitem{chen2013}
Qifeng Chen and Vladlen Koltun.
\newblock {A simple model for intrinsic image decomposition with depth cues}.
\newblock In {\em ICCV}, pages 241--248, 2013.

\bibitem{fan2018}
Qingnan Fan, Jiaolong Yang, Gang Hua, Baoquan Chen, and David Wipf.
\newblock {Revisiting deep intrinsic image decompositions}.
\newblock In {\em CVPR}, pages 8944--8952, 2018.

\bibitem{garces2012}
Elena Garces, Adolfo Munoz, Jorge Lopez-Moreno, and Diego Gutierrez.
\newblock {Intrinsic images by clustering}.
\newblock In {\em Comput. Graph. Forum}, volume~31, pages 1415--1424. Wiley
  Online Library, 2012.

\bibitem{goodfellow2020}
Ian Goodfellow, Jean Pouget-Abadie, Mehdi Mirza, Bing Xu, David Warde-Farley,
  Sherjil Ozair, Aaron Courville, and Yoshua Bengio.
\newblock {Generative adversarial networks}.
\newblock {\em Communications of the ACM}, 63(11):139--144, 2020.

\bibitem{grosse2009}
Roger Grosse, Micah~K Johnson, Edward~H Adelson, and William~T Freeman.
\newblock {Ground truth dataset and baseline evaluations for intrinsic image
  algorithms}.
\newblock In {\em ICCV}, pages 2335--2342. IEEE, 2009.

\bibitem{guislain}
Maximilien Guislain, Julie Digne, Rapha{\"e}lle Chaine, Dimitri Kudelski, and
  Pascal Lefebvre-Albaret.
\newblock {Detecting and correcting shadows in urban point clouds and image
  collections}.
\newblock In {\em 3DV}, pages 537--545. IEEE, 2016.

\bibitem{janner2017}
Michael Janner, Jiajun Wu, Tejas~D Kulkarni, Ilker Yildirim, and Josh
  Tenenbaum.
\newblock {Self-supervised intrinsic image decomposition}.
\newblock {\em NeurIPS}, 30, 2017.

\bibitem{jeon2014}
Junho Jeon, Sunghyun Cho, Xin Tong, and Seungyong Lee.
\newblock {Intrinsic image decomposition using structure-texture separation and
  surface normals}.
\newblock In {\em ECCV}, pages 218--233. Springer, 2014.

\bibitem{kashani2015}
Alireza~G Kashani and Andrew~J Graettinger.
\newblock {Cluster-based roof covering damage detection in ground-based lidar
  data}.
\newblock {\em Automation in Construction}, 58:19--27, 2015.

\bibitem{kingma2013}
Diederik~P Kingma and Max Welling.
\newblock {Auto-encoding variational bayes}.
\newblock {\em arXiv preprint arXiv:1312.6114}, 2013.

\bibitem{krahenbuhl2018}
Philipp Kr{\"a}henb{\"u}hl.
\newblock {Free supervision from video games}.
\newblock In {\em CVPR}, pages 2955--2964, 2018.

\bibitem{land1971}
Edwin~H. Land and John~J. McCann.
\newblock Lightness and retinex theory.
\newblock {\em Journal of the Optical Society of America}, 61(1):1--11, Jan
  1971.

\bibitem{lang2009}
Megan~W Lang and Greg~W McCarty.
\newblock {Lidar intensity for improved detection of inundation below the
  forest canopy}.
\newblock {\em Wetlands}, 29(4):1166--1178, 2009.

\bibitem{larsen2016}
Anders Boesen~Lindbo Larsen, S{\o}ren~Kaae S{\o}nderby, Hugo Larochelle, and
  Ole Winther.
\newblock {Autoencoding beyond pixels using a learned similarity metric}.
\newblock In {\em ICML}, pages 1558--1566. PMLR, 2016.

\bibitem{lee2012}
Kyong~Joon Lee, Qi Zhao, Xin Tong, Minmin Gong, Shahram Izadi, Sang~Uk Lee,
  Ping Tan, and Stephen Lin.
\newblock {Estimation of intrinsic image sequences from image+ depth video}.
\newblock In {\em ECCV}, pages 327--340. Springer, 2012.

\bibitem{Lettry2018}
Louis Lettry, Kenneth Vanhoey, and Luc~Van Gool.
\newblock {Unsupervised Deep Single‐Image Intrinsic Decomposition using
  Illumination‐Varying Image Sequences}.
\newblock {\em Comput. Graph. Forum}, 37, 2018.

\bibitem{li2018cg}
Zhengqi Li and Noah Snavely.
\newblock {Cgintrinsics: Better intrinsic image decomposition through
  physically-based rendering}.
\newblock In {\em ECCV}, pages 371--387, 2018.

\bibitem{li2018}
Zhengqi Li and Noah Snavely.
\newblock {Learning intrinsic image decomposition from watching the world}.
\newblock In {\em CVPR}, pages 9039--9048, 2018.

\bibitem{liao2013}
Zicheng Liao, Jason Rock, Yang Wang, and David Forsyth.
\newblock {Non-parametric filtering for geometric detail extraction and
  material representation}.
\newblock In {\em CVPR}, pages 963--970, 2013.

\bibitem{liu2022}
Xiaoxia Liu, Fengbao Yang, Hong Wei, and Min Gao.
\newblock {Shadow Removal from UAV Images Based on Color and Texture
  Equalization Compensation of Local Homogeneous Regions}.
\newblock {\em Remote Sensing}, 14(11):2616, 2022.

\bibitem{liu2020}
Yunfei Liu, Yu Li, Shaodi You, and Feng Lu.
\newblock {Unsupervised learning for intrinsic image decomposition from a
  single image}.
\newblock In {\em CVPR}, pages 3248--3257, 2020.

\bibitem{luo2020}
Jundan Luo, Zhaoyang Huang, Yijin Li, Xiaowei Zhou, Guofeng Zhang, and Hujun
  Bao.
\newblock {NIID-Net: adapting surface normal knowledge for intrinsic image
  decomposition in indoor scenes}.
\newblock {\em IEEE TVCG}, 26(12):3434--3445, 2020.

\bibitem{ma2018}
Wei-Chiu Ma, Hang Chu, Bolei Zhou, Raquel Urtasun, and Antonio Torralba.
\newblock {Single image intrinsic decomposition without a single intrinsic
  image}.
\newblock In {\em ECCV}, pages 201--217, 2018.

\bibitem{man2015}
Qixia Man, Pinliang Dong, and Huadong Guo.
\newblock {Pixel-and feature-level fusion of hyperspectral and lidar data for
  urban land-use classification}.
\newblock {\em Remote Sensing}, 36(6):1618--1644, 2015.

\bibitem{narihira2015_1}
Takuya Narihira, Michael Maire, and Stella~X Yu.
\newblock {Direct intrinsics: Learning albedo-shading decomposition by
  convolutional regression}.
\newblock In {\em ICCV}, pages 2992--2992, 2015.

\bibitem{narihira2015_2}
Takuya Narihira, Michael Maire, and Stella~X Yu.
\newblock {Learning lightness from human judgement on relative reflectance}.
\newblock In {\em CVPR}, pages 2965--2973, 2015.

\bibitem{nestmeyer2017}
Thomas Nestmeyer and Peter~V Gehler.
\newblock {Reflectance adaptive filtering improves intrinsic image estimation}.
\newblock In {\em CVPR}, pages 6789--6798, 2017.

\bibitem{omer2004}
Ido Omer and Michael Werman.
\newblock {Color lines: Image specific color representation}.
\newblock In {\em CVPR}, volume~2, pages II--II. IEEE, 2004.

\bibitem{priem2016}
Frederik Priem and Frank Canters.
\newblock {Synergistic use of LiDAR and APEX hyperspectral data for
  high-resolution urban land cover mapping}.
\newblock {\em Remote sensing}, 8(10):787, 2016.

\bibitem{rother2011}
Carsten Rother, Martin Kiefel, Lumin Zhang, Bernhard Sch{\"o}lkopf, and Peter
  Gehler.
\newblock {Recovering intrinsic images with a global sparsity prior on
  reflectance}.
\newblock {\em NeurIPS}, 24, 2011.

\bibitem{seo2021}
Kouki Seo, Yuma Kinoshita, and Hitoshi Kiya.
\newblock {Deep Retinex Network for Estimating Illumination Colors with
  Self-Supervised Learning}.
\newblock In {\em LifeTech}, pages 1--5. IEEE, 2021.

\bibitem{shen2011}
Li Shen and Chuohao Yeo.
\newblock {Intrinsic images decomposition using a local and global sparse
  representation of reflectance}.
\newblock In {\em CVPR}, pages 697--704. IEEE, 2011.

\bibitem{ulyanov2018}
Dmitry Ulyanov, Andrea Vedaldi, and Victor Lempitsky.
\newblock {Deep image prior}.
\newblock In {\em CVPR}, pages 9446--9454, 2018.

\bibitem{weiss2001}
Yair Weiss.
\newblock {Deriving intrinsic images from image sequences}.
\newblock In {\em ICCV}, volume~2, pages 68--75. IEEE, 2001.

\bibitem{yu2019}
Ye Yu and William~AP Smith.
\newblock {Inverserendernet: Learning single image inverse rendering}.
\newblock In {\em CVPR}, pages 3155--3164, 2019.

\bibitem{zhang2018}
Yinda Zhang and Thomas Funkhouser.
\newblock {Deep depth completion of a single rgb-d image}.
\newblock In {\em CVPR}, pages 175--185, 2018.

\bibitem{zhao2012}
Qi Zhao, Ping Tan, Qiang Dai, Li Shen, Enhua Wu, and Stephen Lin.
\newblock {A closed-form solution to retinex with nonlocal texture
  constraints}.
\newblock {\em IEEE TPAMI}, 34(7):1437--1444, 2012.

\bibitem{zhou2015}
Tinghui Zhou, Philipp Krahenbuhl, and Alexei~A Efros.
\newblock {Learning data-driven reflectance priors for intrinsic image
  decomposition}.
\newblock In {\em ICCV}, pages 3469--3477, 2015.

\bibitem{zhu2021}
Yongjie Zhu, Jiajun Tang, Si Li, and Boxin Shi.
\newblock {DeRenderNet: Intrinsic Image Decomposition of Urban Scenes with
  Shape-(In) dependent Shading Rendering}.
\newblock pages 1--11. IEEE, 2021.

\end{thebibliography}
}

\end{document}